# IMPROVING CONTENT MARKETING PROCESSES WITH THE APPROACHES BY ARTIFICIAL INTELLIGENCE

**Utku KOSE**
*Usak University - Computer Sciences Application and Research Center, Turkey*
*utku.kose@usak.edu.tr*
**Selcuk SERT**
*Usak University - Karahalli Vocational School*
*selcuk.sert@usak.edu.tr*

**Abstract**
*Content marketing is today's one of the most remarkable approaches in the context of marketing processes of companies. Value of this kind of marketing has improved in time, thanks to the latest developments regarding to computer and communication technologies. Nowadays, especially social media based platforms have a great importance on enabling companies to design multimedia oriented, interactive content. But on the other hand, there is still something more to do for improved content marketing approaches. In this context, objective of this study is to focus on intelligent content marketing, which can be done by using artificial intelligence. Artificial Intelligence is today's one of the most remarkable research fields and it can be used easily as multidisciplinary. So, this study has aimed to discuss about its potential on improving content marketing. In detail, the study has enabled readers' to improve their awareness about the intersection point of content marketing and artificial intelligence. Furthermore, the authors have introduced some example models of intelligent content marketing, which can be achieved by using current Web technologies and artificial intelligence techniques.*

## I. INTRODUCTION

Nowadays, computer and communication technologies have great impact on changing and improving standards of modern life. Since first introduction of especially computer technology, the world has changed fast and greatly in time. It is clear that also advanced communication technologies like Internet have taken technological improvements to a top level, which could not dreamed by people before. Now, almost all fields of the modern life benefit from computer and communication technologies. Marketing is one these fields, which often employ such technologies to improve works and make everything better for more effective and efficient results. As a result of using computer and communication technologies within marketing field, content marketing, which is a new kind of marketing method, has been created. This is another result associated with the evolution of Internet, which has led to the concepts of Internet marketing, digital marketing, social media marketing…etc. (Tantawy and George, 2016). Currently, content marketing is often used for achieving marketing purposes over especially Web platform of Internet.

Because there is a remarkable interest in the idea of that Web and its widely used interactive platforms (i.e. search engines, social media, blogs) are effective on making companies (or brands) more attractive for people (Ahmad et al., 2016; Diaconu et al., 2016; Holliman and Rowley, 2014; Smith and Chaffey, 2013), the Web is preferred by especially successful companies to run their current content marketing processes. Here, delivery and entity of the products are done digitally (Koiso-Kanttila, 2004), and in this way, a more flexible, fast way of marketing is achieved easily. But this type of marketing needs more focus on the elements provided within the marketing processes. That's because even people's roles / behaviors over online platforms are different (Dobre and Milovan-Ciuta, 2015). Hence, there is always a continued need for making even this easy way of marketing better and better to improve whole marketing processes of a company. When we consider the associated literatures, we can see that there are some already done research studies to discuss about potential of relevant business models in making content marketing of some industries more successful (Amit and Zott, 2001; Fetscherin and Knolmayer, 2004; Kaise, 2006; Premkumar, 2003; Swatman et al., 2006; Vaccaro and Cohn, 2004). But if we can also consider multidisciplinary technologies to improve content marketing, it is possible to find out some additional (but also technical), more effective research areas to shape the future of content marketing.

Objective of this study is to focus on intelligent content marketing, which can be done by using artificial intelligence. Artificial Intelligence is today's one of the most remarkable research fields and it can be used easily as multidisciplinary. So, this study aims to discuss about its potential on improving content marketing. In detail, the study also comes with some essential information to enable readers' to improve their awareness about the intersection point of content marketing and artificial intelligence. Additionally, it introduces some example models of intelligent content marketing, which can be achieved by using current Web technologies and artificial intelligence techniques. By focusing on these models, it is aimed to give information about making typical content marketing ways intelligent by using some recent artificial intelligence techniques. All details explained under this study are also important to have idea about future of content marketing, which the authors think that will be done completely computer (artificial intelligence) supported in just near future. From this perspective, the study has a remarkable value of novelty to improve the associated literatures.

The remaining content of the paper / study is organized as follows: The next section is devoted to the concept of content marketing. In detail, this section focuses on some essential information regarding to content marketing and tries to improve readers' knowledge about it. Following to that section, the third section explains the role of artificial intelligence on making content marketing intelligent. In order to achieve that, it discusses about how artificial intelligence can support content marketing and how it can provide effective solutions. After the third section, the fourth section introduces some example models of intelligent content marketing. Under this section, there are a total of six example intelligent content marketing models, which have been designed by the authors. Finally, conclusions and a brief discussion about future work are provided under the last section.

## II. THE CONCEPT OF CONTENT MARKETING

The concept of content marketing can be defined as a marketing approach, which aims to find products produced according to customers' needs and create customer satisfaction and fulfilment in this way (Karkar, 2016). According to the Content Marketing Institute (2016), it is a "strategic marketing approach focused on creating and distributing valuable, relevant, and consistent content to attract and retain a clearly-defined audience - and, ultimately, to drive profitable customer action." As general, a typical content marketing tries to take customers' interest to the products of the company (Ozgen and Doymus, 2013).

### 2.1 Ways of Content Marketing

According to Steimle (2014), the key factor in content marketing is valuable. He says briefly about this subject that "You can tell if a piece of content is the sort that could be part of a content marketing campaign if people seek it out, if people want to consume it, rather than avoiding it" (Steimle, 2014). From a general perspective, it is possible to said that content marketing can be in many different types of formats as follows (De Clerck, 2016; Farnworth, 2015; Omedia24, 2014; Steimle, 2014):
- news,
- video,
- white papers,
- e-books,
- info-graphics,
- e-mail newsletters,
- case studies,
- podcasts,
- how-to guides,
- question - answer articles,
- photos,
- blogs,
- (and many other components, which are enough capable of taking target audience's interest).

### 2.2 Essential Objectives of Content Marketing

The literature comes with many explanations regarding to what are general objectives of a well-formed content marketing. Here, it is possible to focus on Rose and Pullizzi's ideas about objectives of content marketing. They define objectives of content marketing as (Rose and Pullizzi, 2011):
- brand awareness / reinforcement,
- lead conversion and nurturing,
- customer conversion,
- customer service,
- customer upsell,
- passionate subscribers.

All the mentioned objective are tried to be done by designing and developing content marketing models,

thanks to most recent technological factors. At this point, information and communications have an important role on employing some different things to improve / shape present and future state of content marketing. Because of that, the main focus of this study is answering the issue about applying some alternative approaches to improve content marketing and thinking about artificial intelligence for that.

### III. ARTIFICIAL INTELLIGENCE ON CONTENT MARKETING

In order to understand more about the collaboration between content marketing and artificial intelligence, it is better to think about in which aspects of content marketing artificial intelligence can be applied to.

**3.1 Supporting Content Marketing with Artificial Intelligence**
Actually, content marketing can be accepted as a big process including different stages reaching to applying designed content marketing scenario. By eliminating details, it is possible to think about three stages: preparation, application, revision regarding to content marketing. At this point, we can say that the artificial intelligence can be used in any of these stages by choosing appropriate approaches, methods and techniques.

**3.2 Solutions for Content Marketing by Artificial Intelligence**
When we consider the stages we have just indicated for content marketing, it is possible to determine which problem solution ways of artificial intelligence can be used in order to make everything better for content marketing of a company. Some of them are forecasting, optimization, expert support, adaptive guidance (for customers / users), and fixing mistakes (detected along marketing process).

Some popular artificial intelligence techniques regarding to the mentioned solutions / operations are:
- artificial neural networks,
- fuzzy logic,
- genetic algorithms,
- neuro-fuzzy inference systems,
- expert systems,
- swarm intelligence based algorithms,
    - particle swarm optimization,
    - ant colony optimization,
    - artificial bee colony,
    - cuckoo search,
    - krill herd algorithm,
    - differential evolution algorithm,
    - …etc.
- machine learning techniques.

Artificial intelligence field includes many different techniques that are widely used currently. Because the field is very active, newer techniques are often developed. So, in order to have more idea about the field, it is necessary to follow it and gain some pre-knowledge to understand what is actually done. Here, readers interested in learning more about the related techniques are referred to (Alpaydin, 2014; Bolaji et al., 2016; Blum and Li, 2008; Cartwright, 2015; Córdoba et al., 2016; Giarratano and Riley, 1998; Goldberg, 2006; Jackson, 1998; Kennedy et al., 2001; Klir and Yuan, 1995; Kubat, 2015; Mitchell, 1998; Parouha and Das, 2016; Yegnanarayana, 2009).

If we focus on more details regarding to content marketing process, it is possible to figure our more solution ways by artificial intelligence. The main factor about artificial intelligence here is making the content marketing or its process more adaptive, flexible, interactive, and intelligent according to customers' / users' needs and interests. Of course, there are many different ways to achieve this. Also, because there are different kinds of content marketing environments and approaches, it is possible to design many combinations of intelligent content marketing processes.

### IV. EXAMPLE MODELS FOR INTELLIGENT CONTENT MARKETING

In the context of the related subjects that the authors have just focused on, it is possible to think about some example models of intelligent content marketing. Details regarding to six example of models are as follows:

**4.1 Example Model 1: "Intelligent scenario - target customer / user determining"**
First example model of intelligence content marketing is based on dividing the related marketing scenario into some elements and forming a general adaptive scenario changing in time according to feedbacks on the elements. In detail, this scenario is based on a digital content that can be provided over the Web. But the main key factor in this digital content is using some elements to form it. In this context, the digital content may include a media (i.e. video, sound, and animation), some texts, or some using functions that can receive feedback from objective customers / users (i.e. comments under content, like - dislike options, and rating buttons). Here, there is a control mechanism that controls values of each element under an artificial neural networks model and according to the controlled values, a final point for the scenario is determined. Then the final point is used for determining to whom the scenario will be provided. At this point, the target customers / users may be beginner ones, who should see the scenario more, so the scenario is matched with them. On the other hand, it may be possible to show the scenario less to some customers / users with limited elements included under the scenario (It is something like shortening some TV ads after they have been provided for some time). A brief schema regarding to this model is shown under Figure 1.

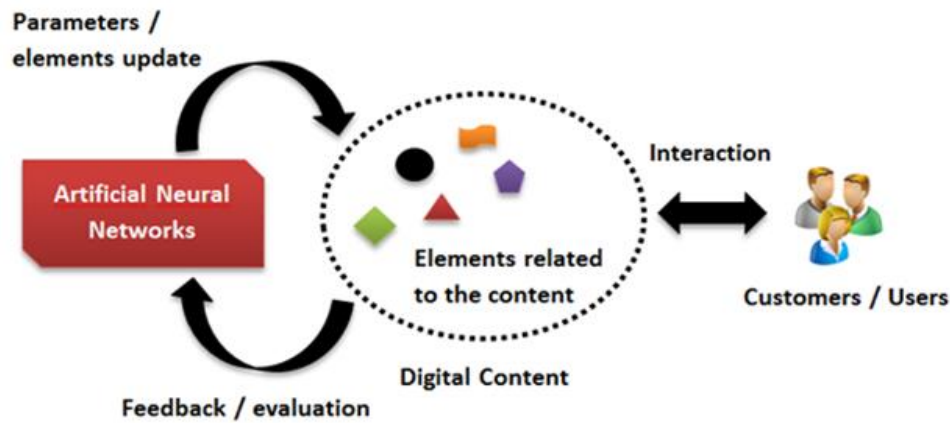

**Figure 1 – Example Model 1: "Intelligent scenario - target customer / user determining".**

### 4.2 Example Model 2: "Optimized scenario"

The second model of intelligent content marketing is related to a typical artificial intelligence based optimization approach. In this model, the scenario (based on digital content) is not evaluated as like it is having some elements. Rather, some key values within the application of the scenario (i.e. number of total views of the scenario, total number of sold, total costs, income, and outcome) have been used for calculating an objective success ratio under a general equation that may be formed by expert(s). In this equation, there are also some additional variables, which should be optimized (i.e. total number of next further view for the scenario, total number for target customers / users, application priority of the scenario among some other alternatives) After a while in which the scenario is applied, the variables are optimized according to objective success ratio and next moves regarding to destiny of the scenario is determined by the people controlling the marketing process, according to new values of the variables. The related processes are continued until the company is happy with the scenario and of course the marketing. As it was pointed under the previous sections, any of artificial intelligence based optimization algorithms can be used for this model. It is also clear that there is a remarkable need for expert support within this model. A brief schema associated with the model - 2 is shown under Figure 2.

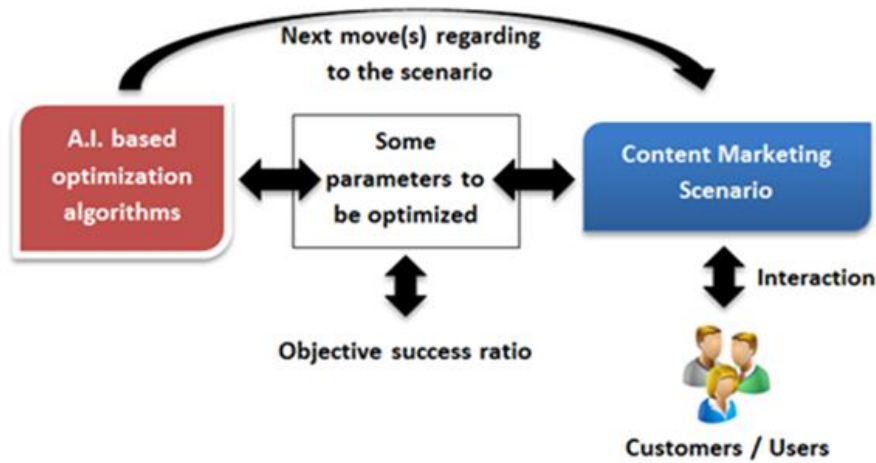

**Figure 2 – Example Model 2: "Optimized scenario".**

**4.3 Example Model 3: "Intelligent evaluation of social media"**

It is clear that social media is a strong environment for companies / brands in order to keep a successful touch with customers. Starting from that point, it is possible to use the social media more for designing some digital content useful for marketing processes. Here, the authors think that the related marketing applications done with the power of social media can be improved by using artificial intelligence. In detail, it is possible to evaluate feedbacks over social media environments and direct social media based operations according to the obtained evaluation results. These evaluation processes may be based on an expert system approach or complex optimization calculations leading to understanding something about i.e. page of the company / brand, and shared ads. Figure 3 presents a brief schema for the model.

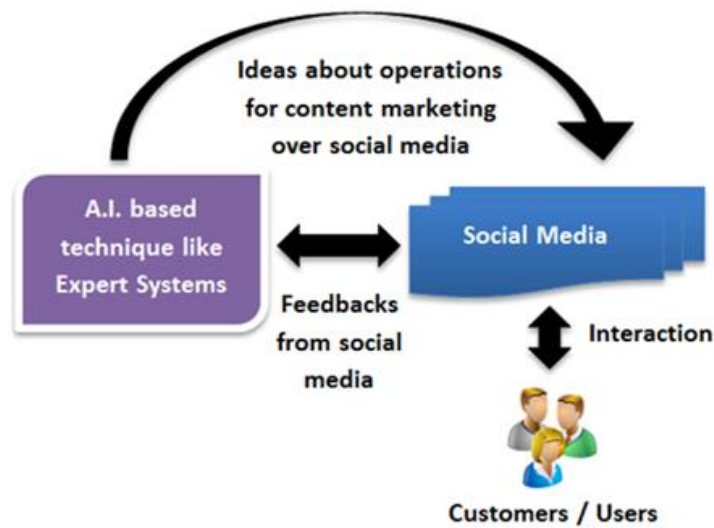

**Figure 3 – Example Model 3: "Intelligent evaluation of social media".**

**4.4 Example Model 4: "Self-learning digital content"**

Another idea of intelligent content marketing is related to use of an intelligent digital content. Main mechanism of such digital content is to update / improve itself by using some parameters related to its possible interactions with customers / users. So, a digital content that have received too low feedback from customers / users may change something in it in order to make itself more attractive or a content showing an increasing popularity may update itself to be adapted more for different Web environments. The intelligent mechanism here can include more than one artificial intelligence technique (like combination of artificial neural networks, and machine learning techniques. Figure 4 shows a brief schema for the model - 4.

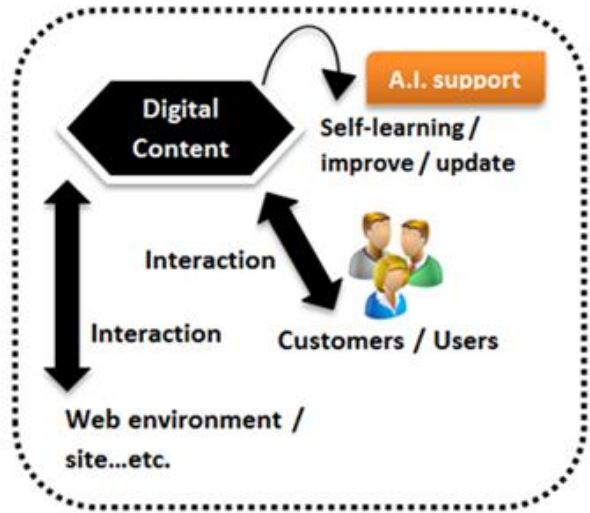

**Figure 4 – Example Model 4: "Self-learning digital content".**

**4.5 Example Model 5: "Intelligent customer / user tracers"**

This model of intelligent content marketing is rather about understanding customers' / users' behaviors over the Web. It may seem something like a spyware but of course, the system on the background of this model includes tracing only customers' / users' actions for some determined points while they are using a Web based platform. In detail, this model includes some kind of swarm, which enables intelligent, small programs to trace customers / users on the background and at the end, inform the main system about which customer / user can be attracted by using which elements over a possible content that may be used for marketing purposes. Briefly this model can be accepted as a typical, intelligent version of cookies used widely over Web platforms that need to store some data for its users. But the approach of intelligent customer / user tracing considered here is more than just storing data. This model is associated with some functions to follow customers / users, gather data while they interact with multimedia objects, Web interfaces, other customers / users…etc., and analyze these data to give some ideas for possible content marketing components (With its final function this model can be even an input to another model, which is able to create content automatically). A brief schema for the model - 5 is shown under Figure 5.

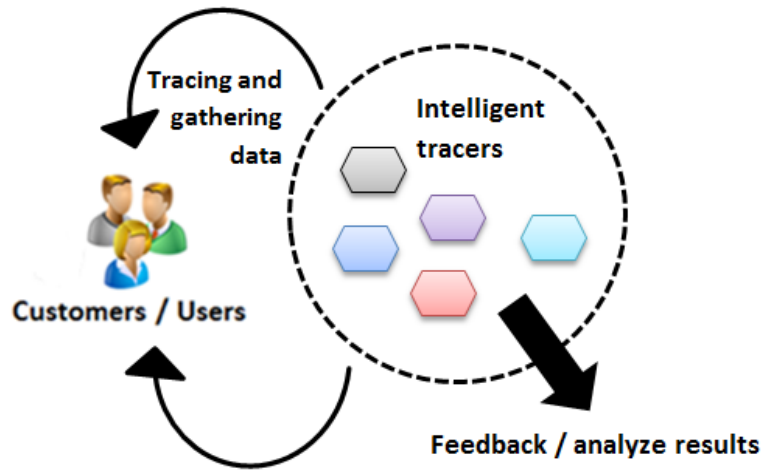

**Figure 5 – Example Model 5: "Intelligent customer / user tracers".**

**4.6 Example Model 6: "Intelligent strategy path determiner"**

It is also possible to design a model for not only directly to the content used for marketing but also for the process done. At this point, a possible way may be to figure out an intelligent system, which is able to determine the most optimum path that companies can use to improve their content marketing processes. As it can be understood, such system needs artificial intelligence techniques to overcome the issues of optimization. But as different from global optimization problems with objective functions, the problem here is designed as a graph, which can be solved by graph-based artificial intelligence optimization algorithms. Each node of the corresponding graph is the move that can be done by the company to improve its content marketing strategy elements. This approach can be done continuously by updating some parameters of the moves after each new process of content marketing, which was performed along a planned time period. In detail, the model considered here is used to receive intelligent suggestions to determine which strategy paths lead the company to be successful at its content marketing processes. Figure 6 presents a brief schema for the model - 6.

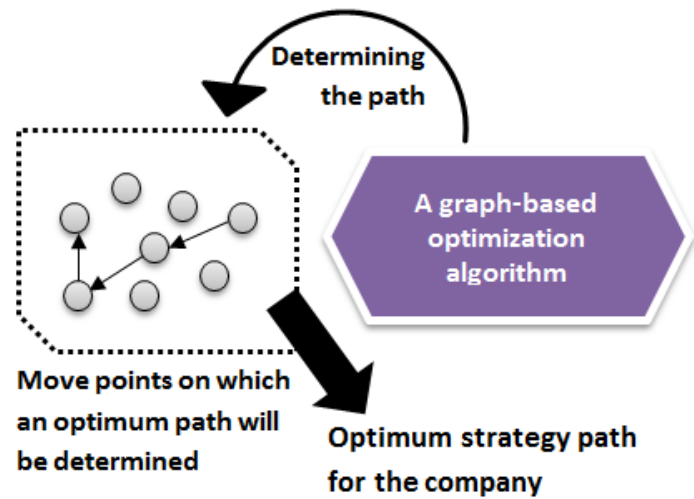

**Figure 6 – Example Model 6: "Intelligent strategy path determiner".**

## V. CONCLUSIONS AND FUTURE WORK

In this study, the concept of intelligent content marketing has been examined briefly. In this context, ideas about using artificial intelligence to improve effectiveness of content marketing have been discussed and they have been supported with some possible example models of intelligent content marketing. In detail, readers are able to have more awareness towards artificial intelligence supported content marketing processes. Furthermore, explanations provided within this study are some kind of ways to understand opportunities about the future of marketing, which lays over content marketing.

As it is also mentioned under the study, models introduced here can be developed by using current artificial intelligence techniques and employing them to popular social media environments or Web platforms. So, future works planned by the authors include developing the related models in a real manner, run them for a while and evaluate obtained results to understand effectiveness and success of intelligent content marketing processes. Additionally, the authors will also continue to work on designing and developing alternative intelligent content marketing models.